\pdfoutput=1

\documentclass[11pt]{article}

\usepackage{ACL2023}

\usepackage{times}
\usepackage{latexsym}
\usepackage{graphicx}
\usepackage{amsmath}
\usepackage{tabularx}
\usepackage{makecell}
\usepackage[T1]{fontenc}

\usepackage[utf8]{inputenc}

\usepackage{microtype}

\usepackage{inconsolata}

\title{Schema-Aware Multi-Task Learning for Complex Text-to-SQL}

\author{Yangjun Wu \\
  Zhejiang University \\
  \texttt{yangjun.wu@connect.polyu.hk} \\\And
  Han Wang \\
  Zhejiang University \\
  \texttt{22021066@zju.edu.cn} \\}

\begin{document}
\maketitle
\begin{abstract}
Conventional text-to-SQL parsers are not good at synthesizing complex SQL queries that involve multiple tables or columns, due to the challenges inherent in identifying the correct schema items and performing accurate alignment between question and schema items. To address the above issue, we present a schema-aware multi-task learning framework (named MTSQL) for complicated SQL queries. Specifically, we design a schema linking discriminator module to distinguish the valid question-schema linkings, which explicitly instructs the encoder by distinctive linking relations to enhance the alignment quality. On the decoder side, we define 6-type relationships to describe the connections between tables and columns (e.g., $WHERE\_TC$), and introduce an operator-centric triple extractor to recognize those associated schema items with the predefined relationship. Also, we establish a rule set of grammar constraints via the predicted triples to filter the proper SQL operators and schema items during the SQL generation. On \textit{Spider}, a cross-domain challenging text-to-SQL benchmark, experimental results indicate that MTSQL is more effective than baselines, especially in extremely hard scenarios. Moreover, further analyses verify that our approach leads to promising improvements for complicated SQL queries.
\end{abstract}

\section{Introduction}
Text-to-SQL aims to automatically translate a natural language utterance to the corresponding executable SQL query in a given database. By helping the great majority of users who are unfamiliar with the query language, this task has many real-world application scenarios and attracts a great deal of interest. Recently, the large-scale cross-domain text-to-SQL dataset \textit{Spider}~\cite{yu-etal-2018-spider} has been released, which contains thousands of complex queries with keywords of JOIN, GROUP BY, etc. Its databases do not overlap between the train and test sets, which requires semantic parsers to have robust inference capability on complex SQL queries.

\begin{figure}
  \centering
  \includegraphics[width=.48\textwidth]{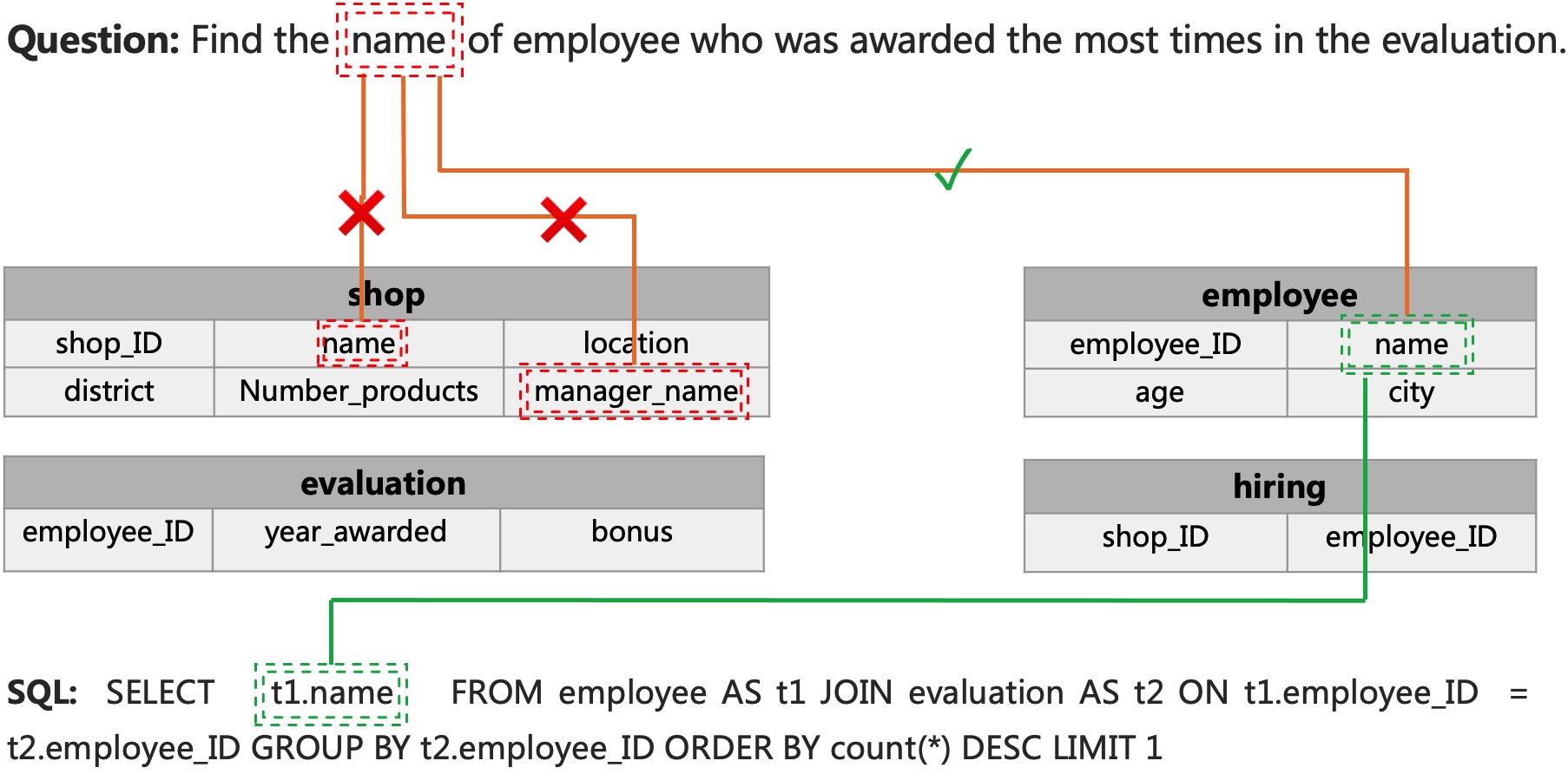}
  \caption{An example from Spider dataset illustrates the existence of valid and invalid alignment to the question word 'name', caused by greedy string matching in the procedure of schema linking.}
  \label{figure1}
\end{figure}

Given a question and a large database with multiple tables, neural semantic parsers are expected to encode the internal database relations and model the alignment between schema items and question words. Most existing works  ~\cite{Guo2019,gan-etal-2021-natural-sql,Rubin2021SmBoPSB,shi2021} inject schema linking into an encoder-decoder network to bridge the mismatch between intent expressed in natural language and the correlated schema table or column. The links are usually obtained via greedy string matching. Considering the example in Figure \ref{figure1}, "$name$" in the question would link to $shop.name$, $employee.name$ and $shop.manager\_name$, but only $employee.name$ is the valid. Unfortunately, all the candidate links are directly fed into the networks to obtain the learnable embeddings regardless of noisy links, where a non-negligible consequence is downgrading the alignment quality and hindering the downstream decoding procedures.

\begin{figure}
  \centering
  \includegraphics[width=.48\textwidth]{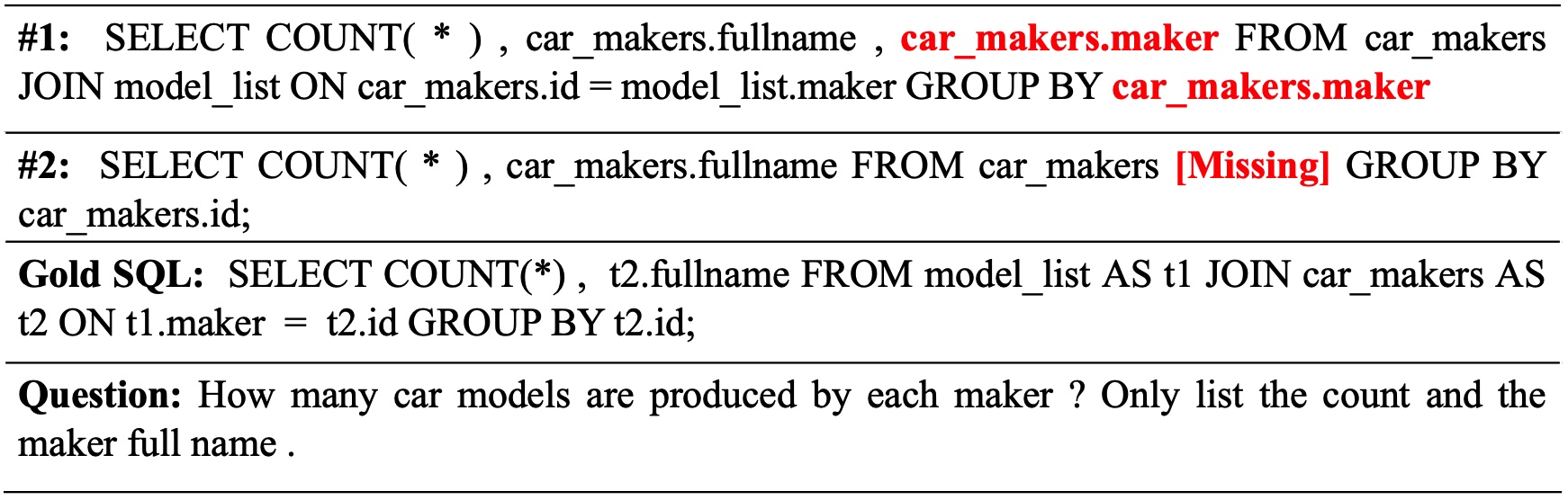}
  \caption{Two error cases to illustrate the obstacles to choosing the correct or complete tables and columns when generating the complicated SQL with JOIN.}
  \label{figure2}
\end{figure}

During the decoding phase, current semantic parsers usually struggle to choose proper schema items (e.g., tables, columns) for those long and hard SQL queries due to the large search space. Moreover, methods that use exact-set-match accuracy \footnote{The definition is available at \url{https://github.com/taoyds/spider/tree/master/evaluation_examples}. This metric decomposes the SQL into several clauses, compares each clause, and returns a boolean value (True or False).} as the evaluation metric would cause the above issue to be more prominent. Because only True or False is given, making it is hard to determine which columns or table names are selected incorrectly. For instance, in the first error case shown in Figure \ref{figure2}, the unexpected column $maker$ in table $car\_makers$ is inferred in the SELECT clause, which makes the clause wrong, and it's not desired in the GROUP BY clause. In another case, SELECT and GROUP BY clauses are correctly generated, but the co-presence schema items ($car\_makers.id = model\_list.maker$) in the JOIN ON clause are failed to reason out. The error cases demonstrate the significance of paying attention to identifying the schema item or co-occurrence schema items while synthesizing complex SQL queries.

Inspired by these observations, we propose a schema-aware \textbf{M}ulti-\textbf{T}ask learning framework called MTSQL to address the two challenging issues. Specifically, we introduce three tasks, including 1) a schema linking classification task to focus on enhancing the quality of alignment between the natural language question and database schemas. 2) A novel phrase-level task named operator-centric triple extraction to concentrate on capturing the correlated schema items with their relationship \footnote{Relationship is operator-centric to represent the links that exist between tables and columns, which include $JOIN\_ON\_TC$, $JOIN\_ON\_CC$, $WHERE\_TC$, $GROUP\_BY\_TC$, $ORDERBY\_TC$, and $SELECT\_TC$. T and C refer to table and column, respectively.}, such as the triple ($car\_makers.\_id$, $model\_list.maker$, $JOIN\_ON\_CC$) extracted from Figure \ref{figure2}. 3) The standard SQL generation task to synthesize complex SQL queries. MTSQL can leverage the fusion of feature information by sharing the weight parameters to raise schema awareness. Further, we utilize the predicted triples to build a rule set as a grammar constraint (GC) module. During the inferencing process, the GC acts as a driver to filter the SQL syntax sub-trees that meet the syntax rules, improving the selection accuracy of SQL operators and schema items.

Experimental results on the benchmark dataset \textit{Spider} indicate that our framework obtains 75.6\% execution with values accuracy on the overall SQL queries, which performs competitive performance. Turning to the complex \footnote{The complex definition is as same as the official \textit{Spider} except that the JOIN ON clause must appear in SQL. The definition is available at \url{https://github.com/taoyds/spider/blob/master/evaluation.py} from the official \textit{Spider}. In short, more complex SQL means more clauses and longer length.} SQL queries with JOIN,  our approach achieves 64.2\% accuracy on the development set with JOIN (\textit{Spider\_join}) and 30.0\% accuracy on another more complex dataset (\textit{United\_Join}), both are state-of-the-art. Particularly, further analyses confirm that MTSQL can lead to 2.4 and 1.6 points improvements on \textit{Spider\_join} and \textit{United\_Join} in extra hard scenarios, which verifies that our proposed framework is more robust and effective for complex text-to-SQL.

\section{Preliminaries}
\subsection{Problem definition}
Given a natural language question $Q$ and the correlated database schema $S$, our goal is to synthesize a corresponding SQL query $Y$. Here, $Q$ = $(w_1,w_2, \ldots, w_m)$ is a sequence of input words. $S$=$(t_1,c_{11},\ldots,t_n,c_{n1}\ldots,c_{nl})$, where $S$ is a database schema. $c_{nl}$ is the {\em l-th} column in the {\em n-th} table $t_n$, which consists of column names and data type $\tau \in\ $\{number, time, text\}. The sequence input $X$ contains the question $Q$ and schema $S$. Two special token $\text{\textless s\textgreater}$ and $\text{\textless /s\textgreater}$ are added to separate the $Q$ and $S$. Here, $X$ is formulated as follows:

\begin{small}
\begin{equation}
\text{\textless s\textgreater},w_1,..,w_m,\text{\textless/s\textgreater},t_1,c_{11},..,t_n,c_{n1}..,c_{nl},\text{\textless /s\textgreater}
\end{equation}
\end{small}The token $\text{\textless s\textgreater}$ at the head of the sequence is used to capture the global contextualized representation.

\begin{table*}[ht!]
\centering
\setlength{\tabcolsep}{3.2mm}{
\begin{tabular}{llll}
\hline
Type of $x$ & Type of $y$ & Relation & Description \\ [1pt]
\hline
table & column & \makecell[l]{tc\_primary\_key \\ tc\_table\_match} &	 \makecell[l]{$y$ is the primary key of $x$. \\ $y$ is a column of $x$ (not primary key).} \\
table	& table &  \makecell[l]{tt\_foreign\_key\_b\\ tt\_foreign\_key\_f} &	\makecell[l]{$y$ and $x$ have foreign keys in both directions. \\ Table $x$ has a foreign key column in $y$.} \\
column	& table & \makecell[l]{ct\_foreign\_key\\ct\_primary\_key} & \makecell[l]{$x$ is the foreign key of $y$.\\ $x$ is the primary key of $y$.} \\
column	& column &\makecell[l]{cc\_table\_match\\
cc\_foreign\_key\\...} & \makecell[l]{$x$ and $y$ belong to same table.\\ $x$ is a foreign key for $y$. \\ ...} \\
\hline
\end{tabular}}
\caption{Description of the internal schema relations. Here, we only list a partial set of relations. 't' and 'c' refer to table and column, respectively.}
\label{tab:rel1}
\end{table*}

\subsection{Schema-aware encoding} For the encoder, there are two types of relationships to be considered, which involve the internal database schema relation (shown in table \ref{tab:rel1}) and the schema linking relation (shown in table \ref{tab:rel2}) between question and schema items. For the former, we follow RAT-SQL\cite{rat-sql} to define dozens of relations, such as {\em tc\_primary\_key} that denotes the column is the primary key of the table. On the schema linking side, the relations contain three categories \{{\em q-tab, q-col, q-value}\} which refer to the alignment between question and table, column, and database cell value, respectively. Each relation is categorized into four cases: \{{\em exact match}, {\em partial match}, {\em stem partial match} and {\em no match}\}.

\begin{table}
\centering
\setlength{\tabcolsep}{0.3mm}{
\begin{tabular}{lllr}
\hline
\small Type of x & \small Type of y &  \small Relation & \small Description \\ [1pt]
\hline
question	& table & \makecell[c]{qt\_partial\_match} &	x is a part of y \\  [1pt]
question	& column &  \makecell[c]{qc\_exact\_match} &	\makecell[r]{x and y are\\ identical} \\[1pt]
question	& value &	 \makecell[c]{qv\_stem\_match \\ ...} & \makecell[r]{x is a part of y \\  after stemming} \\ [1pt]
\hline
\end{tabular}}
\caption{Description of schema linking between question and schema items. To simplify, we just list some types of relations, 'q', 't', 'c', 'v' refer to question, table, column, and cell value, respectively.}
\label{tab:rel2}
\end{table}

\paragraph{Schema-aware self-attention} Relative position representations (RPR) in self-attention \cite{shaw2018-self} is proposed to capture the associated relations among input nodes. In this work, we employ attention mechanism with RPR to capture the relations $R$ that are defined in table \ref{tab:rel1} and \ref{tab:rel2}). The attention computation aims to advance the standard attention procedure and gain schema awareness.

For the attention computation, we first obtain the input embedding $\chi=(h_q;h_s)$ and the relation matrix $R$. $\chi $ contains the question vector $h_q$ and the schema vector $h_s$. $R$ is used to represent the correlated relations among input nodes. $\chi$ and $R$ are fed into the RPR. Then, the distinct attention scores are obtained among nodes. The core attention computation is formulated as follows:

\begin{gather}
e_{i j}^h = \frac{x_i W_Q^h(x_j W_K^h+R_{i j}^K)^\mathrm{T}}{\sqrt{d_{emb} / H}} \\
h_t= attention\left(\chi\right)
\end{gather}

Here, both of $x_i$ and $x_j$ refer to the nodes of input sequence. $R_{ij}^K$ is the learnable relation embedding between $i$ and $j$, and $d_{emb}$ refers to the embedding size. $H$ is the number of heads. Subsequently, we obtain the hidden state of encoder $h_t$ through the novel self-attention mechanism.

\subsection{Bottom-up decoding} \label{bottomup} Current semantic parsers generally synthesize SQL via top-down or bottom-up decoding. In this study, we adopt the identical manner of SmBop\cite{Rubin2021SmBoPSB} to implement a bottom-up abstract syntax tree decoder and utilize the standard query language {\em relational algebra} ~\cite{Codd1970} as the intermediate representation. We recommend the reader to the papers mentioned above for further details.

For bottom-up decoding, we first select the top-K trees of height 0 (leaves), comprising schema items, the conditional values (from utterance), or constants (1, 2, etc.). Then the search procedure starts, and the candidate operators ({\em Selection, Projection, group by, Order by}, etc.) are selected via their scores in the parallel layer by layer. For instance, at step $t$,  the candidate sub-trees ($t+1$ high trees) are constructed from sub-trees with height $t$, and the scores of all the  $t+1$ high sub-trees are computed by the networks. Finally, the tree with the highest score is yielded once the tree's height reaches the maximum value of $T$.

\begin{figure*}[ht]
  \centering
  \includegraphics[width=\textwidth]{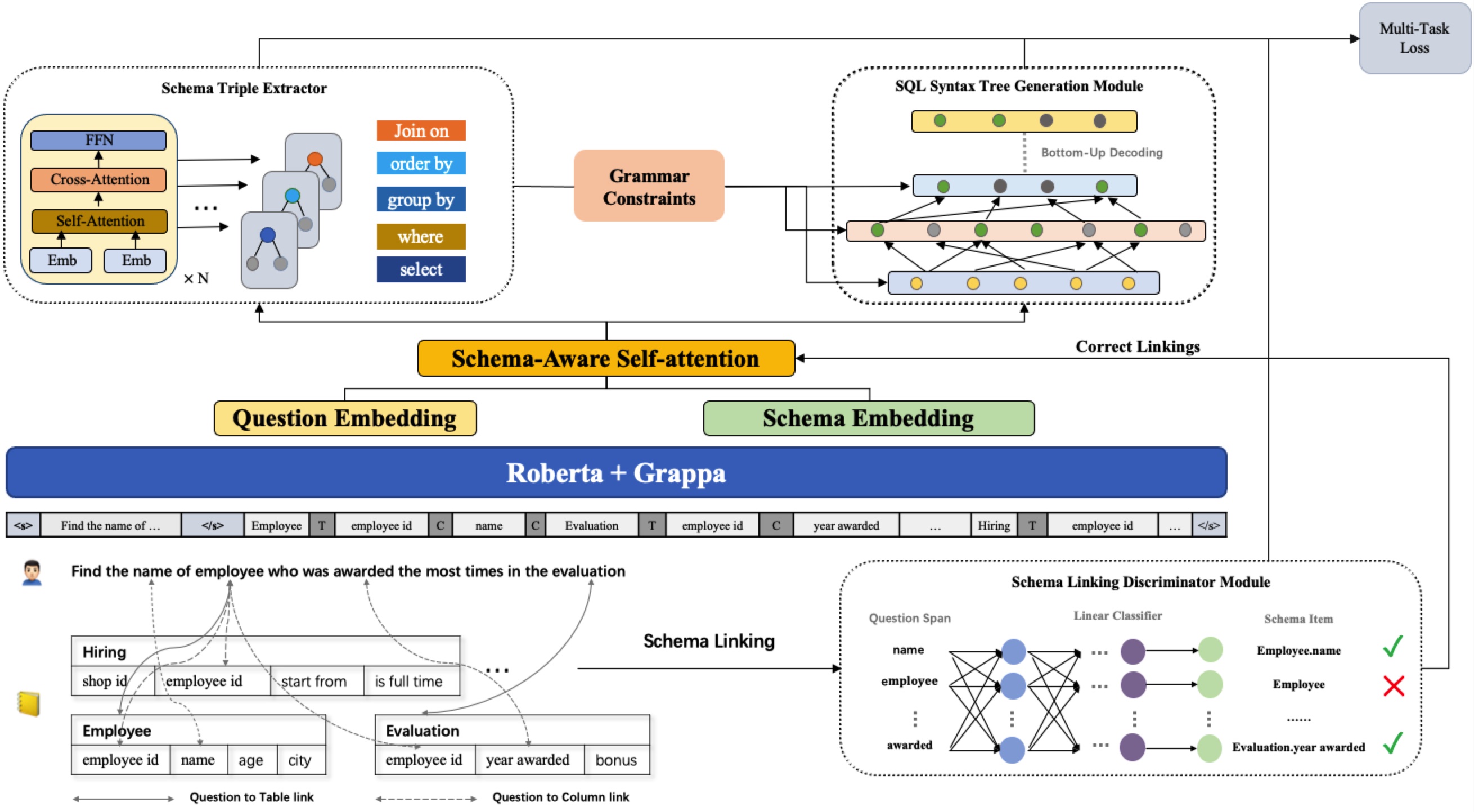}
  \caption{Main MTSQL Architecture. MTSQL contains four core modules: Question-Schema Encoder (QSE), Schema Linking Discriminator module (SLD), Operator-centric Triple Extractor (OTE), SQL Syntax Tree Generation Module (SQLG). In the initialization stage, we apply QSE based on Roberta \cite{Liu2019} with GRAPPA \cite{yu2021grappa} to obtain the joint question-schema feature vector $\chi$. Meanwhile, the correct linkings are filtered by SLD. To better represent the distinct relations among input nodes, the input embedding $\chi$ and the learnable relation matrix $R$ are calculated together during the schema-aware self-attention computation process. At the decoding phase, OTE extracts the significant operator-centric triples (e.g., (table, column, order\_by)). Subsequently, we establish a rule set as grammar constraints by the predicted triples, which drives the bottom-up SQLG to filter the correct SQL operators and schema items. Finally,  the SQL sub-trees are synthesized like beam search, and we produce the top-1 SQL tree as the final SQL.}
  \label{figure3}
\end{figure*}

\section{MTSQL}
\subsection{Overview} 
Figure \ref{figure3} presents the core architecture of MTSQL. Briefly, we design three joint learning tasks for complex text-to-SQL, including schema linking classification task, operator-centric triple extraction, and SQL syntax tree generation task. Besides, the framework is constituted of four modules.

\subsection{Schema Linking Classification} 
This task aims to boost the quality of alignment between the schema items and their mentions in the given question. Therefore, we devise a Schema Linking Discriminator module (SLD) by combining the explicit and implicit methods.

First of all, we utilize greedy string matching to obtain all the preliminary question-schema links. Specifically, for each n-gram from 5 to 1 in the question, we verify whether it exactly matches or is a subsequence of a schema item. If not, we then compare their word stems after stemming ({\em stem partial match}). Otherwise, the relation would be assigned to {\em no match}. Second, all the preliminary links are fed into SLD (based on Multi-Layer Perception MLP) to confirm their validity. The computation procedure is defined as follows:
\begin{gather}
\rho\left(q_i,\ s_j\right) = MLP(q_i , s_j) \\
\rho\left(\theta|q_i,\ s_j\right)=argmax(\rho\left(q_i,\ s_j\right)) \\
 R^i= r^i\ \ if\ \ \rho\left(\theta|q_i,\ s_j\right)\ \geq\ \rho\ ;\ \ i\ \epsilon\ [0,\ M]\label{equa3}
\end{gather} Here, $q_i$ is the {\em i-th} node in the question and  $s_j$ is the {\em j-th} node in the schema. $\rho\left(q,\ s\right)$ refers to the probability distribution in the range of 0 to 1. To guarantee the quality, we only retain those probabilities $\rho\left(\theta|q,\ s\right)$ higher than the threshold $\rho$. $\rho$ is a tunable hyper-parameter. The SLD training object can be formulated as:
\begin{equation}
\begin{aligned}
    \mathcal{L}_{\alpha} = \sum_{s_i}{-\omega_i}[y_i log\rho\left(q_i, s_j\right) \\
    + (1-y_i) log\rho\left(q_i,s_j\right)]
\end{aligned}
\end{equation} Here, we state that the ground truth label of link $y_i$ is 1 if the schema node $s_i$ appears in the gold SQL query.

Finally, SLD selects the accurate relations to form matrix $R$ in Equation \ref{equa3} and feeds $R$ into the schema-aware self-attention for further calculation.

\subsection{Operator-Centric Triple Extraction}
The goal of this task is to select the triples (e.g., (table, column, WHERE\_TC)) that consist of relevant schema items (tables or columns)  with their predefined relationship to the given question. We named them operator-centric triples. 

We treat this task as a set prediction problem and introduce the operator-centric triple extractor (OTE) module based on a non-autoregressive decoder \cite{Gu2018} with Bert \cite{devlin-etal-2019-bert}. Unlike autoregressive approaches, we directly generate all the triples (e.g., {\em (subject, object, relationship)}) in one pass, which demands setting a safe constant $Z$ larger than the maximum number of triples. For the input of the decoder, we randomly initialize $Z$ learnable vectors to represent the triples' embeddings. 

We use the unmasked self-attention at the decoding stage instead of causal mask self-attention to model the relations among triples. Without the constraint of autoregressive factorization (predicting one token at a time from left to right), it can benefit from fully exploiting bidirectional feature information. Then, cross-attention is employed to fuse the information of the question and schema. Finally, the $Z$ vectors are decoded into $Z$ predicted triples by feed-forward networks (FFN). (We add a unique relationship $\oslash$ to refer to the padding triple). For the target triple set $y=\left(s_i^{start},s_i^{end},o_i^{start},o_i^{end},r^i\right)$, where $s_i^{start}$, $s_i^{end}$, $o_i^{start}$, and $o_i^{end}$ are the start or end indices of input sequence, respectively. The conditional probability of the predicted triple set is formulated as follows:
\begin{equation}
\rho\left(\gamma\middle| h_t;\theta\right)= Z \prod_{i=1}^{\kappa}{\rho\left(\gamma_i\middle|h_t,\gamma_{j\neq i};\theta\right)}
\end{equation}
$\rho\left(\gamma_i\middle|h_t,\gamma_{j\neq i};\theta\right)$ denotes the target $\gamma$ is related to $\gamma_{j\neq i}$ and $h_t$ (the hidden state of the encoder). 

\paragraph{Bipartite matching loss} We argue that it's preferable to apply the optimal bipartite matching \cite{Sui2020} (interested readers to the paper for further details) other than cross-entropy as the loss function because it's insensitive to permutations. More precisely, it first finds the optimal matching between the ground truth set $y$ and predicted set $\gamma^\prime$ by Hungarian Algorithm \footnote{https://en.wikipedia.org/wiki/Hungaria\_algorithm} with the lowest expense. The permutation elements of triples $\pi^*$ are formulated as

\begin{equation}
\pi^* = argmin \sum_{i=1}^m{{C_{match}}(\gamma_i, \gamma_i^\prime))}
\end{equation} Here, ${C_{match}}(\gamma_i, \gamma_i^\prime)$ is a pair-wise matching cost between the ground truth $\gamma_i$ and the predicted triple for the index $\pi_i$. The extraction loss $\mathcal{L}_\beta$ is:
\begin{gather}
l^{start} = log{P_{\pi^*(i)}^{start}}{SO_{start}}, SO= [S, O] \\
l^{end} = log{P_{\pi^*(i)}^{end}}{SO_{end}}, SO= [S, O]\\
\mathcal{L}_{\beta} = - \sum_{i=1}^Z {\{ log{P_{\pi^*(i)}^{r}}{R_i}+ l^{start}+ l^{end}\}}
\end{gather} Here, $P_{\pi^*\left(i\right)}^r$ is a probability of the relationship $R_i$, and $i$ ranges in $[1, Z]$. $S_{start}$, $S_{end}$, $O_{start}$, $O_{end}$ are the start or end indices of subject or object. $p_{\pi^*\left(i\right)}^{start}$, $p_{\pi^*\left(i\right)}^{end}$, $p_{\pi^*\left(i\right)}^{start}$, $p_{\pi^*\left(i\right)}^{end}$ are the probabilities of matching between ground truth and the predicted set.

\subsection{SQL Syntax Tree Generation}
The goal of this task is to generate the SQL syntax tree. We follow the work of SmBop and employ bottom-up decoding method (introduced in section \ref{bottomup}) with grammar constraints to synthesize SQL.

\paragraph{Grammar constraints}
Before generating the SQL syntax tree, we exploit the predicted triples to establish a rule set as grammar constraints. At step $t$, the predicted top-K SQL sub-trees have to satisfy the following three grammar rules:
\paragraph{Rule 1.} {\em The leaves of SQL syntax sub-trees must contain the schema items (tables or columns) from the predicted triples by OTE.}
\paragraph{Rule 2.} {\em Increase the scores of the nodes that appeared in the predicted triples.}
\paragraph{Rule 3.} {\em The operators of SQL syntax sub-trees must match the relationship from the predicted triples by OTE.}

During the bottom-up decoding stage, we choose $\frac{K}{2}$ nodes from the candidate schema items and the equivalent nodes from the given question words or constants (1, True, False, etc.) at the first step. Then, we check the top-K nodes with \textbf{Rule 1} for verification. At step $t\ (t> 0)$, the candidate operators ({\em Selection, Order by, etc.}) are predicted in the manner of beam search. Particularly, we employ \textbf{Rule 3} to check the operators and \textbf{Rule 2} to modify operators’ scores. Finally, the syntax tree generator returns the final tree $Y$ with the highest score. Here, we set ${\mathcal{L}}_{\delta}$ as the generation loss.

\paragraph{Multi-Task loss.} The overall weighted loss of MTSQL is:
\begin{equation}
\mathcal{L}\ =\ {{\ \mathcal{L}}_{\delta}\ +\ \lambda\mathcal{L}}_{\alpha}\ +\ {\mu\mathcal{L}}_{\beta}
\end{equation}
Here $\lambda$ and $\mu$ are hyper-parameters.

\section{Experiments}
In this section, we compare the performance of MTSQL with the state-of-the-art approaches on the complex text-to-SQL datasets and further ablate some design choices in MTSQL to understand their contributions.

\subsection{Experiment Setup}
\paragraph{Dataset.}  \textit{Spider} is a popular human-annotated and cross-domain benchmark dataset for text-to-SQL. It contains $7,000$/$1,034$ utterance-SQL pairs in the training/development set and its databases do not overlap between the train and development sets. Since its test set is not publicly accessible, we compare with existing models on the development set (denoted by {\em Spider-Dev}). In this study, we  employ execution accuracy (EX) metric from the official \textit{Spider} to evaluate MTSQL and other approaches. 

Besides \textit{Spider}, we also construct a novel challenging dataset named \textit{United\_Join} to measure the performance on the complex SQL queries. Specifically, we first extract the SQL queries with JOIN operator from the {\em Spider-Dev}  (denoted by {\em Spider\_join}), and the text2sql-data \footnote{The dataset is available at \url{ https://github.com/jkkummerfeld/text2sql-data/}. It's prepared by \cite{FineganDollak2018} and also the recommended dataset by the official \textit{Spider}.} (named {\em text2sql\_join}). Then, we merge  {\em Spider\_join} and {\em text2sql\_join} to form the new dataset \textit{United\_Join}. According to the definition of hardness  level \footnote{\url{https://github.com/taoyds/spider/tree/master/evaluation_examples}} provided by the official \textit{Spider}, \textit{United\_Join} contains $1,442$ question-SQL pairs in total. It consists of $362$ medium, $317$ hard, and $763$ extra-hard samples to the hardness level. Particularly, more than 75\% of the samples in \textit{United\_Join} are beyond the hard level, which the rate is far higher than the proportion of {\em Spider\_join} (32\%). Intuitively, the SQL queries' inference on \textit{United\_Join} is extremely tough and challenging.

\paragraph{Baselines.}  We compare MTSQL with the top performers on the \textit{Spider} Execution leaderboard, including PICARD \cite{Scholak2021}, NatSQL \cite{gan-etal-2021-natural-sql}, SmBop, Bridge\cite{LinRX2020:BRIDGE} and GAZP \cite{zhong2020grounded}. Among these approaches, since bridge trained with text2sql-data, we employ its code to train again on \textit{Spider} for a fair comparison. GAZP only publishes its code without the trained model, thus, we have to train it on our GPU. Because NatSQL is not open-sourced, we can only report NatSQL's results.


\paragraph{Implementations.} We train and test all the models on the $32$GB Tesla V100. Adam optimizer is used with default parameters, and the mini-batch size is set to $30$. In the encoder, we adopt 8-layer schema-aware transformer, and the dropout rate, threshold $\rho$ in Equation \ref{equa3} and the dimension of the hidden unit in SLD are set to $0.2$, $0.995$ and $1024$, respectively. Turning to the SQL syntax tree generator, we utilize the teacher forcing with the dropout rate of $0.5$. We use a 3-layer SQL tree representation to compute the scores, and the top-K and beam size are set to $30$. For OTE, we employ 4-layer transformer-based architecture with $4$ heads. Moreover, the constant $Z$ is set to $20$, and the default weights $\lambda$ and $\mu$ for multi-task overall loss are set to $0.3$ and  $0.05$, respectively.

\begin{table}[ht]
\centering
\setlength{\tabcolsep}{1.4mm}{
\begin{tabular}{lr}
\hline
Model & {\em Spider-Dev} \\[1pt]
\hline
GAZP & 59.2 \\
BRIDGE v1 & 65.3 \\
BRIDGE v2	& 68.0 \\
BRIDGE & 70.3 \\
PICARD + T5-large	& 72.9 \\
NatSQL	& 75.0 \\
SmBoP	& 75.2 \\
\hline
MTSQL	& \textbf{75.6} \\[2pt]
\hline
\end{tabular}}
\caption{Execution accuracy on the \textit{Spider-Dev} set. The highest numbers are in bold.}
\label{tab:tab1}
\end{table}
\subsection{Overall Results}
Table \ref{tab:tab1} shows the performance of MTSQL compared to top performers on the \textit{Spider} Execution leaderboard. We find that MTSQL obtains $75.6\%$ execution accuracy on the {\em Spider-Dev} and outperforms all the other comparison models. The results validate the effectiveness of our proposed scheme-aware multi-task learning framework. 

\begin{table}[ht]
\centering
\setlength{\tabcolsep}{1mm}{
\begin{tabular}{lrrrrr}
\hline
Model & Easy &Medium& Hard& Extra-Hard & All \\[1pt]
\hline
count & 248 & 446 & 174 & 166 & 1034\\[1pt]
\hline
GAZP &	 67.7  & 63.5 &	57.5 &	36.1 &	59.1 \\[1pt]
BRIDGE&	 86.7 &	72.2 &	55.7 & 40.4 & 67.8 \\[1pt]
PICARD & \underline{87.1} &	\underline{74.2} &	\underline{58.0} &	\underline{41.6} &	\underline{69.3} \\[1pt]
SmBoP&	88.7&	78.7&	68.4&	53.0&	75.2 \\
\hline
MTSQL&	86.3&	\textbf{79.4}&	\textbf{70.1}&	\textbf{55.4}&	\textbf{75.6} \\
\hline
\end{tabular}}
\caption{Execution accuracy on the different SQL hardness levels in \textit{Spider\_Dev} set. The underline of numbers indicates that we run the public model to obtain the results. }
\label{tab:tab2}
\end{table}

We also break down the results according to the hardness levels, and their results are shown in Table~\ref{tab:tab2}. Even though MTSQL is slightly inferior to the comparison models when handling Easy problems, it accomplishes the best performance in Medium, Hard, and Extra-Hard levels. We also find that the advantage of MTSQL over the baselines is more prominent while the problems become more complicated. For instance, MTSQL leads to 2.4\% and 3.6\% improvements compared to Smbop and NatSQL in the Extra-Hard conditions. The results are consistent with our motivation to tackle complex text-to-SQL and verify that our proposed techniques are effective for challenging problems.

\begin{table}[ht]
\centering
\setlength{\tabcolsep}{4mm}{
\begin{tabular}{lrr}
\hline
Model&	\textit{Spider\_Join} &	\textit{United\_Join} \\[1pt]
\hline
GAZP & 46.8 &	25.9 \\[1pt]
BRIDGE & 50.5 &	25.6 \\[1pt]
PICARD & 50.2 &	25.4 \\[1pt]
SmBoP &	62 & 27.7\\[1pt]
\hline
MTSQL&	\textbf{64.2}&	\textbf{30.0}\\[1pt]
\hline
\end{tabular}}
\caption{Execution accuracy on \textit{Spider\_Join} and \textit{United\_Join} sets. The  baselines' results are obtained by the public models without modification.}
\label{tab:tab3}
\end{table}

In terms of the complex SQL queries with JOIN, table \ref{tab:tab3} describes the execution accuracy of MTSQL and other baselines. On the {\em Spider\_join} set, MTSQL achieves 64.2\% accuracy, which leads to an increase by 2.3\% over SmBoP while exceeding others by a large margin. Turn to the more challenging set \textit{United\_Join}, MTSQL significantly outperforms all the baselines and achieves the state-of-the-art performance (30\% accuracy). We argue that MTSQL not only pays attention to question-schema alignments but identifies the correct schema items when generating the complicated SQL queries. Therefore, MTSQL increases by 2.3\% and 4.4\% compared to SmBoP and BRIDGE, respectively, and the improvement is in line with our design.

Table \ref{tab:tab4} reports the comparison between MTSQL and the baselines on the \textit{United\_Join} for different hardness level.  MTSQL still directs to 1.0 and 2.7 points improvements compared to SmBoP on the hard and extra-hard levels, which confirms the robustness of MTSQL while predicting the complicated SQL queries with JOIN.


\begin{table}
\centering
\setlength{\tabcolsep}{1.8mm}{
\begin{tabular}{lrrrr}
\hline
Model & Medium& Hard& Extra-Hard & All \\[1pt]
\hline
count&  362 & 317 & 763 & 1442 \\[1pt]
\hline
GAZP &	42.5 &	24.9 & 18.3 &	25.9 \\
BRIDGE & 43.6&	19.2&	19.8&	25.6 \\
PICARD  & 45.3 &	18.9 &	18.6 &	25.4 \\
SmBoP & 49.4&	24.6&	18.7&	27.7 \\
\hline
MTSQL&	\textbf{51.9}&	\textbf{25.6}&	\textbf{21.4}&	\textbf{30.0} \\
\hline
\end{tabular}}
\caption{Execution accuracy on the different SQL hardness levels in \textit{United\_Join}.}
\label{tab:tab4}
\end{table}

\begin{table}[ht]
\centering
\setlength{\tabcolsep}{1.5mm}{
\begin{tabular}{lrr}
\hline
Model&	\textit{Spider-Dev} & \textit{United\_Join} \\
\hline
MTSQL&	75.6&	30.0 \\
- grammar constraints &	75 (-0.6)&	29.6 (-0.4)\\
- SLD&	73.9 (-1.7)&	28.4 (-1.6)\\
- OTE&	72.3 (- 3.3)&	27.4 (-2.6)\\
\hline
\end{tabular}}
\caption{Execution accuracy by removing sub-modules on \textit{Spider-Dev} and \textit{United\_Join}.}
\label{tab:tab5}
\end{table}

\subsection{Ablation study}
We perform a thorough ablation study to show the contribution of each design choice. Table \ref{tab:tab5} and \ref{tab:tab6} are the results of ablation study on the \textit{Spider-Dev} and \textit{United\_Join} sets.

First, we evaluate MTSQL by removing grammar constraints, schema linking discriminator (SLD), or operator-centric triple extractor (OTE). Specifically, the accuracy has a slight drop on the \textit{Spider-Dev} (0.6\%) and \textit{United\_Join} (0.4\%) when withdrawing grammar constraints, which validates that the grammar constraints can effectively restraint the SQL trees generation. The accuracy decreases from 75.6\% to 73.9\% and 30\% to 28.4\% without SLD indicates the necessity of adding one module for high-quality question-schema alignments. Furthermore, the improvements with OTE (3.3\% and 2.6\%) confirm that MTSQL can benefit from the correct schema items selection.

\begin{table}
\centering
\setlength{\tabcolsep}{3.2mm}{
\begin{tabular}{rrrr}
\hline
$\lambda$& $\mu$& \textit{Spider-Dev} & \textit{United\_Join} \\
\hline
0.02&	0.25&	73.5&	28.2\\
0.02&	0.30&	75.2&	29.9\\
0.02&	0.35&	74.6&	29.1\\
0.05&	0.25&	73.9&	28.5\\
0.05&	0.30&	\textbf{75.6}&	\textbf{30.0}\\
0.05&	0.35&	74.8&	29.5\\
\hline
\end{tabular}}
\caption{Execution accuracy by different weight losses hyper-parameters of $\lambda$ and $\mu$.}
\label{tab:tab6}
\end{table}

Second, we further investigate the contribution of different weight loss configurations. Intuitively, the weight parameters can be assigned according to the complexity of sub-tasks, but it's problematic to compute the complexity of the triple sub-tasks. Thus, we tune the hyper-parameters of $\lambda$ and $\mu$ with several manually setting groups. We set the $\lambda$ to 0.02 and 0.05, and $\mu$ to 0.25, 0.3 and 0.35 to estimate the effect, and the results (shown in \ref{tab:tab5}) reveal that the group ($\lambda$ = 0.05,  $\mu$ = 0.30) is the optimal choice. Therefore, we set this group as the default configuration.

\section{Related Works}
\paragraph{Text-to-SQL Semantic Parsing.} The task of Nature Language Interface to Database (NLIDB) has attracted a wide range of interest since the 1970s. In recent years, as several large text-to-SQL datasets are publicly available (e.g., the text2sql-data \cite{FineganDollak2018}, WiKiSQL \cite{zhong2018seqsql}, and \textit{Spider}), plenty of neural semantic parsers have been presented \cite{LinRX2020:BRIDGE,Scholak2021}. The performance in the simple or single-table queries (e.g., WiKiSQL \cite{Xuan2021SeaDET}) is even beyond human beings. However,  the generated results are often not yet satisfactory on the cross-domain text-to-SQL datasets (e.g., \textit{Spider}), especially when addressing the long or extremely complex problems, the performance is still far from the human being.

\paragraph{Schema Linking.}	Schema linking, also known as entity linking, was proposed by \cite{shen2014}. Specifically, entity linking aims to connect entities in a knowledge base with their corresponding mentions in the sentence. It plays a vital role in information extraction, retrieval, etc. In text-to-SQL, the alignment between the natural language utterance and the correlated schema items (tables or columns) is called schema linking, used to capture the unique relations among input nodes. The current approaches can be divided into two groups: explicit and implicit. On the explicit side, we obtain linking relations via string matching in the pre-processing phase, and the relations are directly fed into the encoder. On the other side, the self-attention module \cite{attenion} is employed between question and schema items without the pre-processing. Thus, the linking features are implicitly learned via embedding matching \cite{krishnamurthy2017-neural}. In this work, we incorporate the two types of methods, the initial linking relations are obtained, and then use the SLD to filter the valid relations.

\paragraph{Multi-task Learning (MTL).}	In recent years, MTL has been widely used in pre-training, which aims to boost the performance via multiple interrelated tasks \cite{Caruana1998MultitaskL}. Therefore, it is widely applied in NLP \cite{wang2018-glue}, speech recognition \cite{Deng2013} and computer vision \cite{Zhang2014}, etc. Usually, some parameters among tasks are shared during training, which can benefit from exploring the interrelation among sub-tasks, obtaining extra information, then avoiding overfitting and improving generalization.

\section{Conclusion}
In this paper, we propose a novel schema-aware Multi-Task learning framework called MTSQL for complex text-to-SQL. To augment the  SQL syntax tree generation task, we design a schema linking discriminator module to boost the quality of alignment between natural language question and schema items. We also define 6-type relationships and introduce an operator-centric triple extractor to recognize the related schema items with the predefined relationship. Further, we use the predicted triples to establish a rule set of grammar constraints to filter the accurate SQL operators and schema items during the SQL query generation.

On the \textit{Spider-Dev} and \textit{United\_Join} sets, experimental results demonstrate that our approach is more effective than the baselines while synthesizing the complicated SQL queries. However, the effectiveness is variable under the diverse options of loss weights configurations. Thus, future work can leverage uncertainty \cite{Kendall17} to weight losses other than manually tuning the weighted hyper-parameters to boost the parsing performance.




\bibliography{anthology,custom}

\begin{thebibliography}{27}
\expandafter\ifx\csname natexlab\endcsname\relax\def\natexlab#1{#1}\fi

\bibitem[{Caruana(1998)}]{Caruana1998MultitaskL}
Rich Caruana. 1998.
\newblock Multitask learning.
\newblock In \emph{Encyclopedia of Machine Learning and Data Mining}.

\bibitem[{Codd(1970)}]{Codd1970}
E.~F. Codd. 1970.
\newblock \href {https://doi.org/10.1145/362384.362685} {A relational model of data for large shared data banks}.
\newblock \emph{Commun.ACM}.

\bibitem[{Deng et~al.(2013)Deng, Hinton, and Kingsbury}]{Deng2013}
Li~Deng, Geoffrey Hinton, and Brian Kingsbury. 2013.
\newblock \href {https://doi.org/10.1109/ICASSP.2013.6639344} {New types of deep neural network learning for speech recognition and related applications: an overview}.
\newblock In \emph{ICASSP}.

\bibitem[{Devlin et~al.(2019)Devlin, Chang, Lee, and Toutanova}]{devlin-etal-2019-bert}
Jacob Devlin, Ming-Wei Chang, Kenton Lee, and Kristina Toutanova. 2019.
\newblock \href {https://doi.org/10.18653/v1/N19-1423} {{BERT}: Pre-training of deep bidirectional transformers for language understanding}.
\newblock In \emph{Proceedings of the 2019 Conference of the North {A}merican Chapter of the Association for Computational Linguistics: Human Language Technologies, Volume 1 (Long and Short Papers)}, pages 4171--4186, Minneapolis, Minnesota. Association for Computational Linguistics.

\bibitem[{Finegan-Dollak et~al.(2018)Finegan-Dollak, Kummerfeld, Zhang, Ramanathan, Sadasivam, Zhang, and Radev}]{FineganDollak2018}
Catherine Finegan-Dollak, Jonathan~K. Kummerfeld, Li~Zhang, Karthik Ramanathan, Sesh Sadasivam, Rui Zhang, and Dragomir~R. Radev. 2018.
\newblock Improving text-to-sql evaluation methodology.
\newblock \emph{ArXiv}, abs/1806.09029.

\bibitem[{Gan et~al.(2021)Gan, Chen, Xie, Purver, Woodward, Drake, and Zhang}]{gan-etal-2021-natural-sql}
Yujian Gan, Xinyun Chen, Jinxia Xie, Matthew Purver, John~R. Woodward, John Drake, and Qiaofu Zhang. 2021.
\newblock \href {https://doi.org/10.18653/v1/2021.findings-emnlp.174} {Natural {SQL}: Making {SQL} easier to infer from natural language specifications}.
\newblock In \emph{Findings of the Association for Computational Linguistics: EMNLP 2021}, pages 2030--2042, Punta Cana, Dominican Republic. Association for Computational Linguistics.

\bibitem[{Gu et~al.(2017)Gu, Bradbury, Xiong, Li, and Socher}]{Gu2018}
Jiatao Gu, James Bradbury, Caiming Xiong, Victor O.~K. Li, and Richard Socher. 2017.
\newblock \href {http://arxiv.org/abs/1711.02281} {Non-autoregressive neural machine translation}.
\newblock \emph{CoRR}, abs/1711.02281.

\bibitem[{Guo et~al.(2019)Guo, Zhan, Gao, Xiao, Lou, Liu, and Zhang}]{Guo2019}
Jiaqi Guo, Zecheng Zhan, Yan Gao, Yan Xiao, Jian-Guang Lou, Ting Liu, and Dongmei Zhang. 2019.
\newblock \href {https://doi.org/10.18653/v1/P19-1444} {Towards complex text-to-{SQL} in cross-domain database with intermediate representation}.
\newblock In \emph{Proceedings of the 57th Annual Meeting of the Association for Computational Linguistics}, pages 4524--4535, Florence, Italy. Association for Computational Linguistics.

\bibitem[{Kendall et~al.(2017)Kendall, Gal, and Cipolla}]{Kendall17}
Alex Kendall, Yarin Gal, and Roberto Cipolla. 2017.
\newblock \href {http://arxiv.org/abs/1705.07115} {Multi-task learning using uncertainty to weigh losses for scene geometry and semantics}.
\newblock \emph{CoRR}, abs/1705.07115.

\bibitem[{Krishnamurthy et~al.(2017)Krishnamurthy, Dasigi, and Gardner}]{krishnamurthy2017-neural}
Jayant Krishnamurthy, Pradeep Dasigi, and Matt Gardner. 2017.
\newblock \href {https://doi.org/10.18653/v1/D17-1160} {Neural semantic parsing with type constraints for semi-structured tables}.
\newblock In \emph{EMNLP}.

\bibitem[{Lin et~al.(2020)Lin, Socher, and Xiong}]{LinRX2020:BRIDGE}
Xi~Victoria Lin, Richard Socher, and Caiming Xiong. 2020.
\newblock Bridging textual and tabular data for cross-domain text-to-sql semantic parsing.
\newblock In \emph{Findings, {EMNLP}}.

\bibitem[{Liu et~al.(2019)Liu, Ott, Goyal, Du, Joshi, Chen, Levy, Lewis, Zettlemoyer, and Stoyanov}]{Liu2019}
Yinhan Liu, Myle Ott, Naman Goyal, Jingfei Du, Mandar Joshi, Danqi Chen, Omer Levy, Mike Lewis, Luke Zettlemoyer, and Veselin Stoyanov. 2019.
\newblock \href {http://arxiv.org/abs/1907.11692} {Roberta: {A} robustly optimized {BERT} pretraining approach}.
\newblock \emph{CoRR}, abs/1907.11692.

\bibitem[{Rubin and Berant(2021)}]{Rubin2021SmBoPSB}
Ohad Rubin and Jonathan Berant. 2021.
\newblock \href {https://doi.org/10.18653/v1/2021.naacl-main.29} {{S}m{B}o{P}: Semi-autoregressive bottom-up semantic parsing}.
\newblock In \emph{Proceedings of the 2021 Conference of the North American Chapter of the Association for Computational Linguistics: Human Language Technologies}, pages 311--324, Online. Association for Computational Linguistics.

\bibitem[{Scholak et~al.(2021)Scholak, Schucher, and Bahdanau}]{Scholak2021}
Torsten Scholak, Nathan Schucher, and Dzmitry Bahdanau. 2021.
\newblock \href {https://doi.org/10.18653/v1/2021.emnlp-main.779} {{PICARD}: Parsing incrementally for constrained auto-regressive decoding from language models}.
\newblock In \emph{Proceedings of the 2021 Conference on Empirical Methods in Natural Language Processing}, pages 9895--9901, Online and Punta Cana, Dominican Republic. Association for Computational Linguistics.

\bibitem[{Shaw et~al.(2018)Shaw, Uszkoreit, and Vaswani}]{shaw2018-self}
Peter Shaw, Jakob Uszkoreit, and Ashish Vaswani. 2018.
\newblock \href {https://aclanthology.org/N18-2074} {Self-attention with relative position representations}.
\newblock In \emph{NAACL-HLT(Short Papers)}.

\bibitem[{Shen et~al.(2015)Shen, Wang, and Han}]{shen2014}
Wei Shen, Jianyong Wang, and Jiawei Han. 2015.
\newblock \href {https://doi.org/10.1109/TKDE.2014.2327028} {Entity linking with a knowledge base: Issues, techniques, and solutions}.
\newblock \emph{TKDE}.

\bibitem[{Shi et~al.(2020)Shi, Ng, Wang, Zhu, Li, Wang, dos Santos, and Xiang}]{shi2021}
Peng Shi, Patrick Ng, Zhiguo Wang, Henghui Zhu, Alexander~Hanbo Li, Jun Wang, C{\'{\i}}cero~Nogueira dos Santos, and Bing Xiang. 2020.
\newblock \href {http://arxiv.org/abs/2012.10309} {Learning contextual representations for semantic parsing with generation-augmented pre-training}.
\newblock \emph{CoRR}, abs/2012.10309.

\bibitem[{Sui et~al.(2020)Sui, Chen, Liu, Zhao, Zeng, and Liu}]{Sui2020}
Dianbo Sui, Yubo Chen, Kang Liu, Jun Zhao, Xiangrong Zeng, and Shengping Liu. 2020.
\newblock \href {http://arxiv.org/abs/2011.01675} {Joint entity and relation extraction with set prediction networks}.
\newblock \emph{CoRR}, abs/2011.01675.

\bibitem[{Vaswani et~al.(2017)Vaswani, Shazeer, Parmar, Uszkoreit, Jones, Gomez, Kaiser, and Polosukhin}]{attenion}
Ashish Vaswani, Noam Shazeer, Niki Parmar, Jakob Uszkoreit, Llion Jones, Aidan~N. Gomez, \L{}ukasz Kaiser, and Illia Polosukhin. 2017.
\newblock Attention is all you need.
\newblock In \emph{Proceedings of the 31st International Conference on Neural Information Processing Systems}, NIPS'17, page 6000–6010, Red Hook, NY, USA. Curran Associates Inc.

\bibitem[{Wang et~al.(2018)Wang, Singh, Michael, Hill, Levy, and Bowman}]{wang2018-glue}
Alex Wang, Amanpreet Singh, Julian Michael, Felix Hill, Omer Levy, and Samuel Bowman. 2018.
\newblock \href {https://doi.org/10.18653/v1/W18-5446} {{GLUE}: A multi-task benchmark and analysis platform for natural language understanding}.
\newblock In \emph{Proceedings of the 2018 {EMNLP} Workshop {B}lackbox{NLP}: Analyzing and Interpreting Neural Networks for {NLP}}.

\bibitem[{Wang et~al.(2020)Wang, Shin, Liu, Polozov, and Richardson}]{rat-sql}
Bailin Wang, Richard Shin, Xiaodong Liu, Oleksandr Polozov, and Matthew Richardson. 2020.
\newblock {RAT-SQL}: Relation-aware schema encoding and linking for text-to-{SQL} parsers.
\newblock In \emph{ACL}.

\bibitem[{Xuan et~al.(2021)Xuan, Wang, Wang, Wen, and Dong}]{Xuan2021SeaDET}
Kuan Xuan, Yongbo Wang, Yongliang Wang, Zujie Wen, and Yang Dong. 2021.
\newblock Sead: End-to-end text-to-sql generation with schema-aware denoising.
\newblock \emph{ArXiv}, abs/2105.07911.

\bibitem[{Yu et~al.(2021)Yu, Wu, Lin, bailin wang, Tan, Yang, Radev, richard socher, and Xiong}]{yu2021grappa}
Tao Yu, Chien-Sheng Wu, Xi~Victoria Lin, bailin wang, Yi~Chern Tan, Xinyi Yang, Dragomir Radev, richard socher, and Caiming Xiong. 2021.
\newblock \href {https://openreview.net/forum?id=kyaIeYj4zZ} {Grappa: Grammar-augmented pre-training for table semantic parsing}.
\newblock In \emph{ICLR}.

\bibitem[{Yu et~al.(2018)Yu, Zhang, Yang, Yasunaga, Wang, Li, Ma, Li, Yao, Roman, Zhang, and Radev}]{yu-etal-2018-spider}
Tao Yu, Rui Zhang, Kai Yang, Michihiro Yasunaga, Dongxu Wang, Zifan Li, James Ma, Irene Li, Qingning Yao, Shanelle Roman, Zilin Zhang, and Dragomir Radev. 2018.
\newblock \href {https://doi.org/10.18653/v1/D18-1425} {{S}pider: A large-scale human-labeled dataset for complex and cross-domain semantic parsing and text-to-{SQL} task}.
\newblock In \emph{Proceedings of the 2018 Conference on Empirical Methods in Natural Language Processing}, pages 3911--3921, Brussels, Belgium. Association for Computational Linguistics.

\bibitem[{Z. et~al.(2014)Z., C.C., and X.}]{Zhang2014}
Zhang Z., Luo P.and~Loy C.C., and Tang X. 2014.
\newblock Facial landmark detection by deep multi-task learning.
\newblock \emph{ECCV}.

\bibitem[{Zhong et~al.(2020)Zhong, Lewis, Wang, and Zettlemoyer}]{zhong2020grounded}
Victor Zhong, Mike Lewis, Sida~I. Wang, and Luke Zettlemoyer. 2020.
\newblock Grounded adaptation for zero-shot executable semantic parsing.
\newblock In \emph{EMNLP}.

\bibitem[{Zhong et~al.(2018)Zhong, Xiong, and Socher}]{zhong2018seqsql}
Victor Zhong, Caiming Xiong, and Richard Socher. 2018.
\newblock \href {https://openreview.net/forum?id=Syx6bz-Ab} {Seq2{SQL}: Generating structured queries from natural language using reinforcement learning}.

\end{thebibliography}
\bibliographystyle{acl_natbib}




\end{document}